# An Adaptive ML Framework for Power Converter Monitoring via Federated Transfer Learning

Panagiotis Kakosimos, *Senior Member, IEEE*, Alireza Nemat Saberi, and Luca Peretti, *Senior Member, IEEE*

*Abstract*—This study explores alternative framework configurations for adapting thermal machine learning (ML) models for power converters by combining transfer learning (TL) and federated learning (FL) in a piecewise manner. This approach inherently addresses challenges such as varying operating conditions, data sharing limitations, and security implications. The framework starts with a base model that is incrementally adapted by multiple clients via adapting three state-of-the-art domain adaptation techniques: Fine-tuning, Transfer Component Analysis (TCA), and Deep Domain Adaptation (DDA). The Flower framework is employed for FL, using Federated Averaging for aggregation. Validation with field data demonstrates that fine-tuning offers a straightforward TL approach with high accuracy, making it suitable for practical applications. Benchmarking results reveal a comprehensive comparison of these methods, showcasing their respective strengths and weaknesses when applied in different scenarios. Locally hosted FL enhances performance when data aggregation is not feasible, while cloud-based FL becomes more practical with a significant increase in the number of clients, addressing scalability and connectivity challenges.

## I. INTRODUCTION

THE growing demand for advanced condition monitoring has increased the adoption of machine learning (ML) techniques. Although traditional threshold-based approaches are commonly used, they have shown limitations in effectively monitoring the dynamic characteristics of modern power systems [1]. Additionally, analytical models may be unavailable for complex systems, or when they do exist, they often lack the scalability and adaptability required for monitoring large numbers of assets autonomously, without human intervention. In contrast, data-driven approaches can learn from data and make accurate predictions while adapting to unseen scenarios to some extent [2].

However, as ML models become more integrated into industrial processes, challenges arise in handling collected data and generalizing developed models. Universal ML models could be one solution if assets shared common characteristics and operating ranges, but this is not always possible. Transfer learning (TL) has emerged as a technique to leverage information from multiple assets [3]. In [4], a deep convolutional TL network was adapted to learn domain-invariant features for condition monitoring purposes when labeled data are scarce. In cases where distribution discrepancies exist across machines, regularization terms are incorporated in the training process to impose constraints on the model parameters and reduce the distribution mismatch between the source and target domains [5]. Instead of adapting the model in the training process, feature representation across domains can be discovered and then used to transform the data [6]. In [7], a method to learn some transfer components was developed while preserving the data properties and underlying relationships. Therefore, with the representations in the new subspace spanned by these components, standard ML training methods can be used.

For established TL approaches to be applied, sharing data and model configuration between domains is often required. However, this is not always possible due to privacy challenges. Additionally, as the number of devices grows rapidly, there is a strong need for benefiting from collective knowledge [8], [9]. Federated Learning (FL) enables collaborative model training by exchanging model properties instead of raw data [10], [11]. In [12], a Gaussian mixture model for feature sharing and a dynamic weighted aggregation algorithm were presented to address distribution differences in battery state-of-health estimation. Similarly, an attention-free transformer coupled with a weighted aggregation method was presented in [13] for combining run-to-failure data from multiple assets. While FL helps collaboratively train ML models, data privacy concerns also increase due to several attack mechanisms [14].

In this paper, we present an adaptive framework that integrates transfer learning (TL) and federated learning (FL) in a sequential manner to enhance the monitoring of power converters under diverse operating conditions. The approach starts with a base model that is incrementally adapted by multiple clients using three of the most widely recognized as state-of-the-art domain adaptation techniques: Fine-tuning, Transfer Component Analysis (TCA), and Deep Domain Adaptation (DDA). Afterward, the clients are trained in a collaborative manner orchestrated by the FL server to evaluate possible network configurations and contribute to the training of a global model. Through validation with field data, we demonstrate that fine-tuning achieves high accuracy, making it practical for real-world applications. The comprehensive comparison results demonstrate that locally hosted FL boosts

Panagiotis Kakosimos is with the Corporate Research Center, ABB AB, 72358 Västerås, Sweden (e-mail: Panagiotis.Kakosimos@se.abb.com).

Alireza Nemat Saberi is with the Motion System Drives, ABB Oy, 00380 Helsinki, Finland (e-mail: alireza.nemat-saberi@fi.abb.com).

Luca Peretti is with the Division of Electric Power and Energy Systems, KTH Royal Institute of Technology, 10044 Stockholm, Sweden (e-mail: lucap@kth.se).

performance when data aggregation is impractical, while cloud-based FL offers scalability benefits as the number of clients increases. These benchmarking results highlight the significance of choosing the appropriate framework based on specific needs and conditions of power converters.

## II. METHODOLOGY

The primary goal of this study is to investigate the application of federated transfer learning (FTL) in the estimation of internal temperatures in electric drives with ML. The emphasis is on the implementation of FTL in various settings and its potential to improve model adaptation across different environments. More specifically, various ML approaches have been examined in [1] for thermal modeling of power converters, with the Multilayer Perceptron (MLP) emerging as a suitable candidate due to its balance between accuracy and computational efficiency, thus it has been adopted in this study. The input features for the ML model include current, ambient temperature, frequency, and power, with the output feature being the relative temperature, defined as the difference between the power module and ambient temperatures. The use of the relative temperature inherently accounts for variations in ambient temperature, allowing the model to focus on learning patterns related to internal heat generation and dissipation. Furthermore, this approach facilitates transfer learning, as the model becomes less dependent on specific ambient conditions and more focused on intrinsic thermal behavior. Detailed information on the base ML model development, parameter selection, and performance is available in [1].

Given that power converters operate on similar thermal principles but are deployed in different environments with varying operating conditions, a tailored FTL approach is needed. When the converters' operating conditions differ substantially, it is possible to achieve implicit TL through the global model within the FL in exchange for additional training rounds. However, considering that some converters may experience extreme or unique conditions in an uncontrolled manner, the performance at the local level will be impacted. Additionally, neural network models tend to experience catastrophic forgetting, where models continuously overfit their most recent training data [15]. In the context of FL, the phenomenon is expected to be exacerbated due to the heterogeneous nature of the data [11]. Therefore, in this study, the models of each client are firstly adapted via TL and then integrated into the FL process, as illustrated in Fig. 1. The investigation is also conducted sequentially, beginning with testing and evaluating various TL approaches to assess their suitability and integrability within the FTL framework. Subsequently, the implementation of FL with the selected TL approach is validated across different scenarios to evaluate the performance.

### A. Transfer learning

Although the clients correspond with similar asset types, their exposure to different operating conditions and design variations necessitates the use of a TL approach to maintain client-size adjustments in a reversible manner. Therefore, in this study, TL is implemented locally by evaluating three broadly used methods, after adjusting them for power converters, in order to achieve the best compromise between complexity and accuracy [16]: (1) Fine-tuning, (2) Transfer Component Analysis (TCA), and (3) Deep Domain Adaptation (DDA).

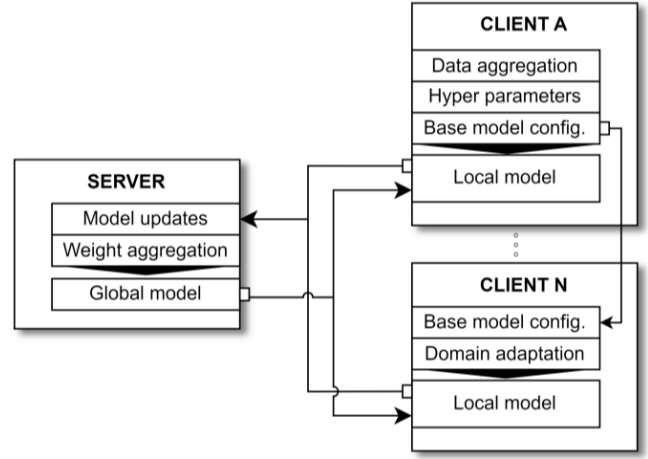

**Fig. 1.** Overview of the FTL framework.

Fine-tuning is the most straightforward technique in which the layers of a pre-trained base model are frozen, and only the newly added layers are trained using a portion of the target dataset [17]. This approach allows the model to leverage knowledge from the pre-trained layers while adapting to the specific features of the new data. As shown in Fig. 2, fine-tuning does not involve any direct data adaptation, as the task-specific adjustments are entirely managed by the layer adjustments.

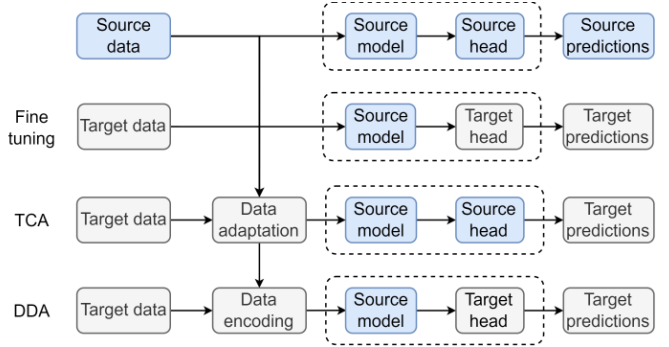

**Fig. 2.** Overview of the three investigated TL approaches.

On the other hand, domain adaptation is a more specialized approach that focuses on aligning the distributions of the source and target domains even with substantial divergence. Domain adaptation almost universally utilizes Maximum Mean Discrepancy (MMD) as an unsupervised feature alignment metric, especially among state-of-the-art methods. The MMD between the two distributions, $x^S$ and $x^T$, is defined as [18]:

$$\text{MMD}(x^S, x^T) = \| \frac{1}{N^S} \sum_{i=1}^{N^S} \phi(x_i^S) - \frac{1}{N^T} \sum_{i=1}^{N^T} \phi(x_i^T) \|_{\mathcal{H}}^2, \quad (1)$$

where $x^S$ and $x^T$ are samples from the two domains, $N^S$ and $N^T$ are the numbers of samples, and $\|\cdot\|_{\mathcal{H}}^2$ denotes the two-norm operation in a reproducing kernel Hilbert space. A Gaussian kernel, $\phi$, is commonly used as the mapping function, since it weighs closer points more, helping in capturing local data patterns.

In this study, MMD has been used differently in TCA and DDA. In TCA, MMD is used to encode the source and target datasets, accounting for domain variations, as shown in Fig. 3 [7]. The main reason for using MMD in TCA is to transform the input data into a domain-invariant feature space locally on each client rather than changing the model structure and weights. This enables the integration of the derived model into the FL process without further changes.

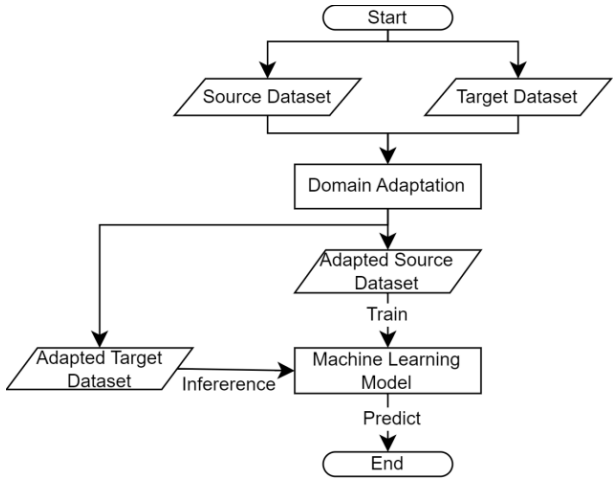

**Fig. 3.** Flow diagram of the TCA methodology.

On the other hand, in DDA, MMD is incorporated in the loss function along with the Mean Squared Error (MSE) from a frozen decoder and a weight regularization term, as depicted in Fig. 4 [4], [5]. Specifically, a pre-trained model is first developed by training an autoencoder on the source data, after which the decoder is frozen. By simultaneously optimizing the combined loss functions, DDA aims to achieve an optimal balance between maintaining the source domain's characteristics, aligning the source and target domains, and enhancing the model's ability to generalize effectively. Since the model structure is not expected to differ among the clients, DDA can be incorporated in the FL architecture.

*B. Distributed training*

Initiating model training in a distributed learning setup introduces additional complexities, particularly in scenarios with restricted data sharing. This study examines and benchmarks almost all possible combinations, ranging from fully centralized to completely isolated configurations. The FTL framework begins with a first client (i.e., Client A) initiating the training process. After short-term data aggregation, hyperparameters are selected, and a base model is generated. This base model is then shared with subsequent clients, who join the training process sequentially. Once the clients are configured with the base model configuration (Fig. 2), the distribution of training resources begins with the FL setup. Given that the thermal models of the power converters share the same feature space but differ in operational data, a horizontal FL structure is implemented to facilitate the FTL process. In horizontal FL, different clients hold distinct samples of data that share the same features. This contrasts with vertical FL, where different clients possess different features but the same samples, making it suitable for scenarios where assets have complementary data.

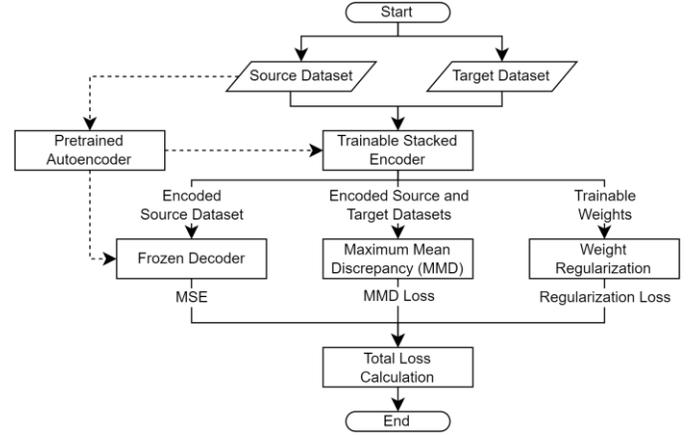

**Fig. 4.** Deep domain adaptation flowchart.

To orchestrate the learning process, the Flower framework, widely used in the literature, has been utilized [19]. In this arrangement, the clients transmit their local model updates to the FL server, which can be hosted either locally or in the cloud. Among the available aggregation techniques, Federated Averaging (FedAvg) stands out for its performance, thus it has also been adopted in this study [20]. More specifically, the FL server aggregates the updates to adjust the global model before redistributing it to the clients by using a weighted average formula:

$$\mathbf{w}_{t+1} = arg\ min \sum_{k=1}^{K} \frac{n_k}{n} \mathcal{L}_k(\mathbf{w}), \qquad (2)$$

where $\mathbf{w}$ represents the global model parameters, $\mathcal{L}_k(\mathbf{w})$ is the local loss function for client $k$, $n$ is the number of data samples, and $K$ is the total number of clients. In (2), the weights of each client are allocated based on the size of the dataset each client uses for training (i.e., $w_k = \frac{n_k}{n}$).

Since the local models either have structural changes or require data transformations, their integration with the aggregation technique exhibits differences (Fig. 2). In fine-tuning, the base model layers are frozen during TL, whereas during FL, it is the turn of the previously trainable layers to freeze their weights. Thus, the aggregation process is realized via partial model averaging on the FL updated layers only. In contrast, TCA learns the underlying common components across clients' data distributions without directly updating the model's weights but by applying data transformations. In DDA, the model structure is maintained in the same way as TCA by both loss function changes and data transformations; however, its complexity increases significantly as can be seen in Fig. 4.

To balance the performance of FL against changes in the working conditions of the clients, a more dynamic method of weight allocation may be used. In such a scenario, the metric of MMD can adjust the weights, $w_k = \frac{n_k \cdot R_k}{\sum_{j=1}^{N} n_j \cdot R_j}$, where $R_k$ is the relevance score of client $k$. The limitation of this approach is that the global model may become excessively influenced by clients with working conditions similar to the majority, reducing the contributions from clients with distinct or less common conditions. This may lead to suboptimal performance for those outlier clients and limit the overall model's ability to generalize across diverse scenarios. Therefore, in this study, the FL framework is combined with a TL strategy.

Although FL reduces the need for data sharing, frequent exchange of model information may result in overhead communication and delays. On the other hand, increasing the number of training rounds helps improve accuracy but intensifies the communication burden, thus necessitating a balance to justify using FL and to determine the most suitable server location. Therefore, this study comprehensively benchmarks centralized training and isolated clients alongside local and cloud-based FL implementations within the FTL framework, as shown in in Fig. 5.

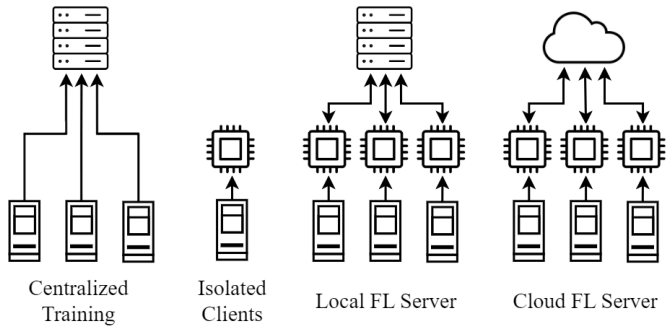

**Fig. 5.** Investigated network topologies integrating FL.

## III. RESULTS & DISCUSSION

The various aspects of the TL and FL approaches investigated within the proposed framework have been analyzed by using real-world data collected in electric drives. Fig. 6 shows a representative converter environment of a pumping station, an application responsible for approximately a quarter of global energy consumption. The used converter topology is the conventional two-level; however, the results can be applied to any converter topology. The power module temperature is measured at the baseplate using an integrated NTC thermocouple, available in most commercial power modules (Fig. 6) [21], [22]. Despite that the sensor accuracy is around ±1°C, the thermal ML model is utilized to estimate the sensor output, focusing on detecting unexpected variations rather than monitoring absolute temperature values [1].

### A. Transfer learning

The probability density functions of the source and one of the target domains as shown in Fig. 7 exhibit significantly different shapes and cover dissimilar ranges across the four input features, necessitating the use of TL. For the mapping of the source and target domains, the domain with sufficient data was designated as the source, while other domains with limited data served as the targets.

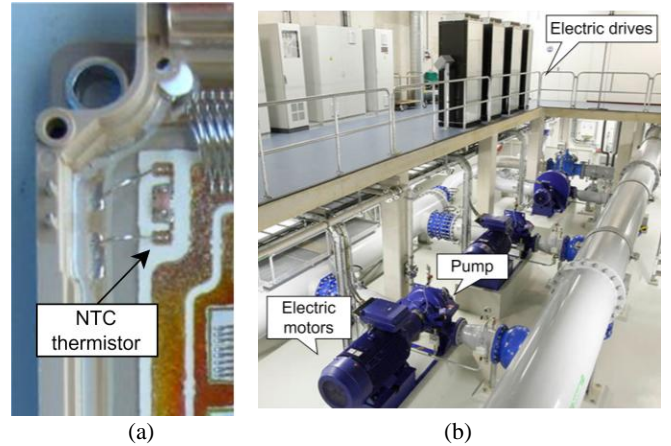

**Fig. 6.** (a) Example of the position of the integrated NTC thermistor [22] and (b) representative converter environment (source:abb.com).

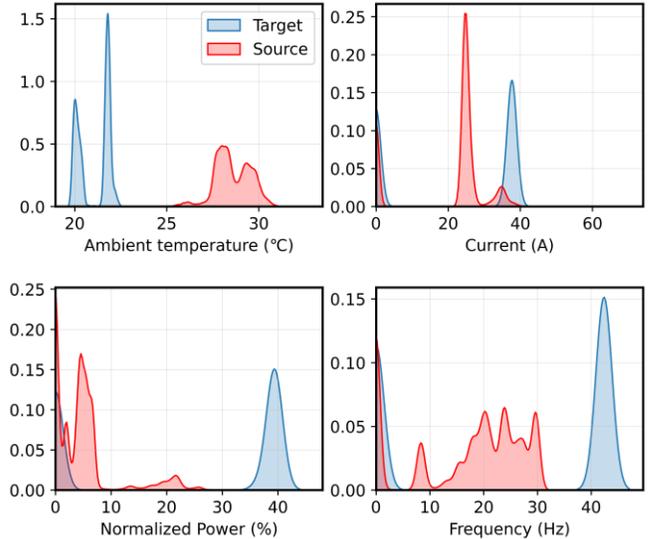

**Fig. 7.** Probability density functions of the source and target datasets.

The temporal variation of the input features for the target domain is shown in Fig. 8. The field data used reflect typical real-world operational conditions, where intermittent operation is common, thus causing temperature swings. In many cases, drives are over-dimensioned relative to the application needs, leading to partial loading. These conditions present significant challenges, as threshold-based temperature monitoring approaches may fail to trigger alerts, even in cases of issues such as cooling inefficiencies. Additionally, the ambient temperature measured at the air inlet of the drive is slightly influenced by the drive's operation due to their proximity. Given that the main objective is to monitor temperature increases on top of the ambient temperature, the use of relative temperature is further justified.

Figures 9 and 10 compare the performance of the three evaluated TL methods, illustrating the measured versus predicted relative temperatures. Both fine-tuning and DDA show central error tendencies close to zero, while TCA exhibits slight bias. Nevertheless, all methods demonstrate satisfactory performance with minimal variability. Since the drive operates dynamically with frequent starts and stops, the temperature error exhibits spikes (Fig. 9). The use of a 1-minute sampling interval intensifies the discrepancy due to artificial mismatches between the measurement and prediction. Nevertheless, the estimation error falls within the accuracy range of temperature sensors.

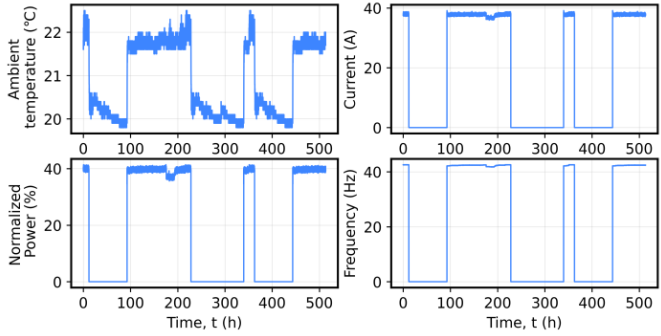

**Fig. 8.** Input features used for the target domain, shown for an input signal subset of Fig. 7.

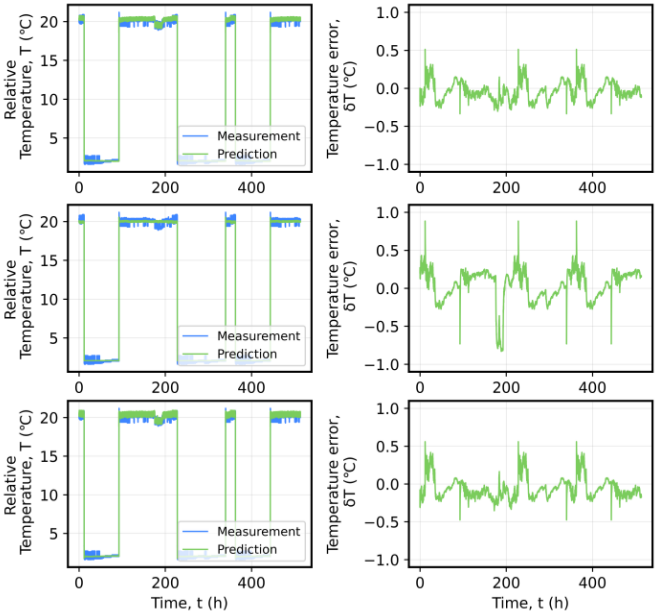

**Fig. 9.** Measured and predicted relative temperatures for fine-tuning, TCA, and DDA, from top to bottom, shown for an input signal subset of Fig. 8.

From an implementation perspective within the framework, fine-tuning has the lowest complexity, as it relies on an existing base model with adaptation of the newly added trainable layers. In contrast, TCA and DDA require the alignment of feature distributions with data transformation and encoding before model training or inference. This renders these methods more robust when there is no abundance of target data or when the resource usage must be adjusted dynamically.

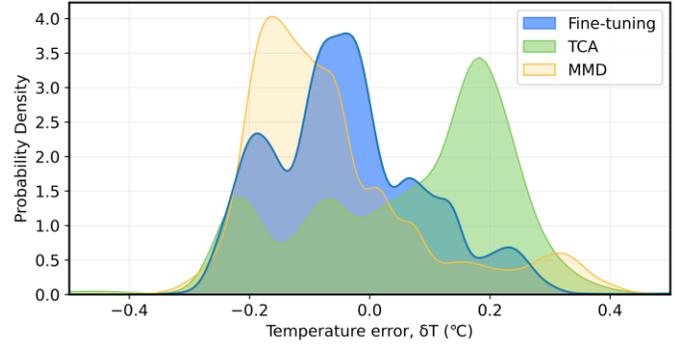

**Fig. 10.** Temperature error probability densities for all methods from Fig. 9.

The three state-of-the-art methods, as they have been adjusted for the context of power converters, are compared in Table I in both target and source domains. As discussed previously, all methods demonstrate low errors in the target domain even without further optimization of the model hyperparameters. However, when the re-trained models are evaluated on the adapted source data, the errors increase for all methods, with fine-tuning showing the largest error. This can be attributed to the lower variability in the target data distribution, as illustrated in Fig. 7, which contributes to improved performance in the target domain. This scenario is commonly observed in real-world or field settings, where power converters may function under both static and dynamic conditions. Consequently, choosing the most appropriate method depends on the specific application and its requirements. Given that TL is combined with FL in the investigated framework, fine-tuning has been selected due to its lower implementation complexity while maintaining sufficient accuracy.

TABLE I
PERFORMANCE COMPARISON OF TL APPROACHES

| Method | Target domain | | Source domain | |
|---|---|---|---|---|
| | MSE | $R^2$ | MSE | $R^2$ |
| Fine-tuning | 0.095 | 0.992 | 7.543 | 0.821 |
| TCA | 0.163 | 0.998 | 6.921 | 0.892 |
| DDA | 0.105 | 0.996 | 3.625 | 0.939 |

Afterward, the fine-tuned model is tested under larger temperature variations, with increased power and current of approximately 25%. To accomplish this, larger load variations are incorporated into the measurements, which are then used to adapt the target through fine-tuning. The fine-tuned target model is then inferred using additional operating conditions, as depicted in Fig. 11. The operating conditions in Fig. 8 were synthetically adjusted by adapting the current and power for the range from 300 to 400 hours of operation to test the model under this new scenario. Despite the increase in temperature, the model's behavior is not significantly affected by the new operating conditions. Provided that the training dataset is sufficient, or the signal relationships remain valid for the model to extrapolate, when necessary, its behavior is expected to remain consistent. As anticipated, the error of the relative temperature increases immediately at the load change but remains at low values.

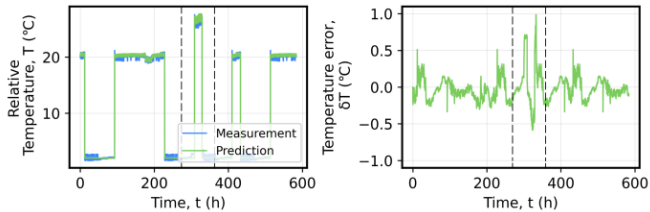

**Fig. 11.** Adjusted measured and predicted relative temperatures for fine-tuning, shown for an input signal subset of Fig. 8 where the current and power have been synthetically increased by 25%.

*B. Distributed learning*

After establishing a local model for each client by adapting the base model, the distributed learning process is initiated within the framework (Fig. 1). A global model is then generated on the FL server, which coordinates the clients' training using locally collected data. Even though a base model was initially used for transfer learning and to generate the global model, all clients contribute periodic updates based on their individual data distributions. As the FL process continues, the local models will start to deviate from the global one, which, in turn, will reflect a combination of all clients due to the weight aggregation.

The benchmarking results of all combinations within the framework is summarized in Fig. 12. As expected, centralized training with aggregated data achieves the lowest error, while training clients independently in a siloed environment results in suboptimal performance with noticeable variability. Hosting the FL server locally, even with a limited number of 10 training rounds, achieved similar results. Increasing the number of rounds to 100 led to performance improvements, but it still did not reach that of centralized training. The scenario involving a cloud server produced different results than the other configurations. This discrepancy was primarily due to intermittent connection losses with some clients. After exceeding a certain number of efforts, the server continues with the clients with an established connection.

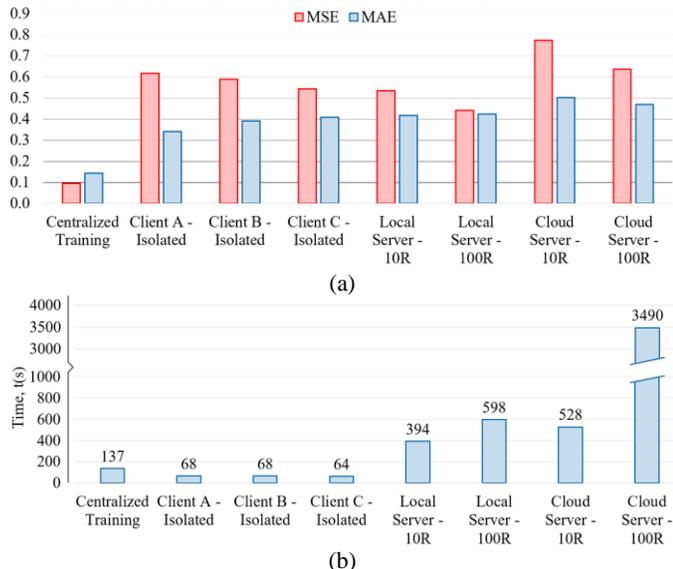

**Fig. 12.** Model performance (a) and training times (b) for the examined scenarios of Fig. 5.

Increasing the number of rounds further enhanced the cloud-based FL performance, but a tenfold increase in the number of rounds resulted in nearly a sevenfold increase in training time. To accommodate this, the number of local epochs was reduced. The results suggest that FL offers improved and more consistent performance when hosted locally. However, this does not address the challenges of isolated systems, and data aggregation remains a promising alternative. Hosting the FL server in the cloud becomes viable when the number of clients is significantly larger, reducing the impact of connectivity issues and making data aggregation impractical due to the high storage demands.

Although FL improved model performance compared to isolated training, individual client models in the FTL framework are still susceptible to catastrophic forgetting, where previously learned information is lost when training on new data. In the investigated cases, it is not possible to separate one client's data into fully independent and non-identically distributed subsets of previous and new data (Fig. 8). Hence, to evaluate this effect, an additional experiment was conducted, where clients were fine-tuned at the final stage of FL using their initial dataset. The results showed that model performance improved by 1.5–1.8% after re-fine-tuning, indicating that some previously learned knowledge was not fully retained during FL training. To mitigate this effect, storing and selectively reintroducing past data batches for periodic fine-tuning could be considered, helping to reinforce critical information and enhance long-term model performance.

## V. CONCLUSION

This study examines and benchmarks various up-to-date configurations for adapting and generalizing thermal ML models for power converters in electric drives, with a primary focus on the integration of TL and FL. The stepwise integration within the proposed framework inherently addresses the challenges of varying operating conditions and restricted data sharing. Fine-tuning emerged as the most suitable TL method for edge deployment, owing to its lower complexity and satisfactory performance. The distributed training process, orchestrated by the Flower framework, effectively coordinated learning across multiple clients. Local hosting of the FL server demonstrated improved performance and consistency, whereas cloud-based FL exhibited potential scalability benefits when the number of clients increased significantly. The FTL framework offers a scalable solution for condition monitoring in real-world environments, ensuring accuracy and minimal data transmission. Nevertheless, optimizing the balance between training rounds and communication overhead remains crucial for maximizing the efficiency and effectiveness of the FTL approach.